\documentclass[twoside]{article}

\usepackage[accepted]{aistats2020}

\usepackage{hyperref}
\usepackage{AVTcommands}
\usepackage{microtype}

\usepackage[bibstyle=trad-abbrv, citestyle=authoryear, backend=bibtex, backref=true, useprefix, maxcitenames=2]{biblatex}
\DeclareFieldFormat{titlecase}{#1}
\renewbibmacro*{pageref}{%
  \iflistundef{pageref}
    {}
    {\addperiod\addspace\printtext{%
      \ifnumgreater{\value{pageref}}{1}
         {\bibstring{backrefpages}\ppspace}
         {\bibstring{backrefpage}\ppspace}%
      \printlist[pageref][-\value{listtotal}]{pageref}}}}

\renewcommand{\multicitedelim}{\addsemicolon\space}
\DeclareCiteCommand{\cite}[\mkbibparens]{\usebibmacro{prenote}}{\usebibmacro{citeindex}\usebibmacro{cite}}{\multicitedelim}{\usebibmacro{postnote}}
\defbibenvironment{bibliography}
  {\list
     {}
     {\setlength{\leftmargin}{\bibhang}%
      \setlength{\itemindent}{-\leftmargin}%
      \setlength{\itemsep}{\bibitemsep}%
      \setlength{\parsep}{\bibparsep-0.5ex}}}
  {\endlist}
  {\item}
\usepackage{xpatch}
\xpatchbibmacro{cite}{\printnames{labelname}}{\printtext[bibhyperref]{\printnames{labelname}}}{}{}
\xpatchbibmacro{textcite}{\printnames{labelname}}{\printtext[bibhyperref]{\printnames{labelname}}}{}{}

\bibliography{references}
\let\UrlFont\scshape

\usepackage{subfig}

\renewcommand{\[}{\begin{equation}}
\renewcommand{\]}{\end{equation}}
\def\<#1\>{\begin{align}#1\end{align}}
\def\?#1\?{\begin{gather}#1\end{gather}}

\begin{document}

\runningtitle{Variational Integrator Networks for Physically Structured Embeddings}

\twocolumn[
\aistatstitle{Variational Integrator Networks\\for Physically Structured Embeddings}
\aistatsauthor{Steind\'{o}r S{\ae}mundsson \And Alexander Terenin \And Katja Hofmann\And Marc Peter Deisenroth}
\aistatsaddress{Imperial College London \And Imperial College London \And Microsoft Research \And University College London}
]

\begin{abstract}
Learning workable representations of dynamical systems is becoming an increasingly important problem in a number of application areas.
By leveraging recent work connecting deep neural networks to systems of differential equations, we propose \emph{variational integrator networks}, a class of neural network architectures designed to preserve the geometric structure of physical systems. 
This class of network architectures facilitates accurate long-term prediction, interpretability, and data-efficient learning, while still remaining highly flexible and capable of modeling complex behavior.
We demonstrate that they can accurately learn dynamical systems from both noisy observations in phase space and from image pixels within which the unknown dynamics are embedded.
\end{abstract}

\section{Introduction} \label{sec:intro}

Deep learning has revolutionized application areas, such as image classification and reinforcement learning, in part via its ability to obtain representations of data that generalize well and are useful for downstream tasks.
Deep networks have accomplished this by simultaneously being highly expressive, yet capable of learning effectively from a finite amount of data.
A key determinant in this efficiency is the inductive bias encoded by the architecture of the network, such as in convolutional networks for image data, as well as long short-term memory networks for text and other sequential data.
These structural assumptions allow the network to learn efficiently, while still enabling it to capture complex relationships that are prohibitively difficult to feature engineer or write down manually.

We are interested in applying such networks to dynamical systems governed by the laws of physics.
Such systems are highly flexible and capable of modeling complex phenomena. However, they also possess inherent structure, such as conservation laws.
In machine learning, this important structure is often ignored, due to the black-box nature of off-the-shelf algorithms.
To perform well on a given task, deep neural networks must learn to conserve these quantities as effectively as possible.
Owing to the precise form of their equations, such networks generally do not conserve these quantities exactly \cite{Greydanus2019}.
 \textcite{Greydanus2019} demonstrated that this flaw harms the networks' capacity for accurate long-term prediction.

As a workaround, \textcite{Greydanus2019} proposed to parameterize the dynamical system's Hamiltonian using a neural network, and to learn it directly from data.
The specification of the Hamiltonian fully determines the dynamics. The equations of motion are then reconstructed from the learned Hamiltonian via standard techniques from mechanics.
One downside to this approach is the black-box nature of the neural network, which makes it difficult to encode properties of the dynamical system, such as its constraints or symmetries. \textcite{Lutter2019} propose an architecture that imposes Lagrangian mechanics, and is optimized to minimize the violation of the equations of motion. A similar idea is also used in \cite{Raissi2019} to learn general non-linear differential equations from physics. A potential drawback of encoding physical plausibility through the loss function is the need for training data that reasonably covers the configuration space. 

The continuous-time equations of motion for a dynamical system are given by a set of differential equations that can be derived from its Lagrangian via variational calculus.
These equations encode the underlying physical properties, such as conservation laws.
In parallel, a deep residual network can be viewed as an Euler discretization of a system of ordinary differential equations, see \textcite{Haber2017,E2017,Chen2018}.

In this paper, we aim to bridge the viewpoint of neural ODEs \cite{Haber2017,E2017,Chen2018,Chang2018, Ruthotto2018}, where neural networks are seen as discretized dynamical systems, with the viewpoint of geometric embeddings \cite{Chamberlain2017, Nickel2017, Ganea2018, Davidson2018}, which impose structure on an embedding space.
When data is concentrated on a manifold, \textcite{Falorsi2018} argued that it is crucial to ensure the embedding space has the same topology as this manifold, motivating Lie group variational auto-encoders \cite{Falorsi2018, Haan2018, Falorsi2019}.

We propose to model the dynamical system using a deep neural network, whose architecture matches the discrete-time equations of motion governing the dynamical system.
This allows us to re-interpret the embedding learned by the network as a dynamical system in its own right.
We focus on a class of discretization methods called \emph{variational integration}~\cite{Marsden2001}.
This gives rise to our proposed \emph{variational integrator networks}:
a class of flexible neural network architectures that encode physical laws and manifold constraints by preserving the underlying geometry inherent to physical systems. These properties promote accurate long-term prediction, interpretability and more efficient learning than is possible with comparable black-box function approximators.

We demonstrate their effectiveness on a number of tasks, including inferring dynamical systems from noisy observations, and from the pixels of images, both in an interpretable and data-efficient manner\footnote{Code available on GitHub: \url{https://github.com/steindoringi/Variational_Integrator_Networks}}.
\section{Variational Integrators}
\begin{figure*}
\begin{center}
\includegraphics{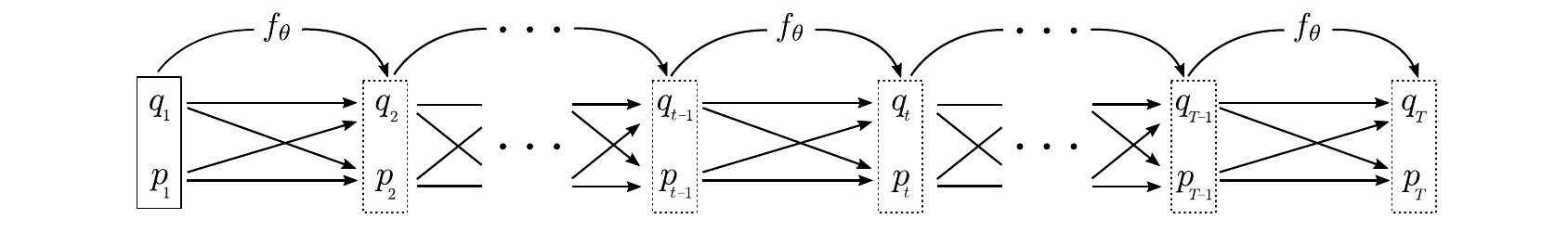}
\end{center}
\caption{Variational integrator network. Here, $(\v{q},\v{p})$ are the hidden states, and $f_\theta$ is a residual block. The full variational integrator network is built by stacking free-form residual blocks in the manner prescribed by a variational integrator to obtain a deep network.} \label{fig:vins}
\end{figure*}

In this section, we review variational integrators (VIs), a general class of discretization methods for dynamical systems.
We study physical dynamical systems over a configuration space $\c{Q}$, with generalized positions and velocities denoted by $\v{q}(t),\dot{\v{q}}(t)$. The systems are governed by the principle of least action, specified via the Lagrangian $L(\v{q}(t),\dot{\v{q}}(t))$, and expressible in Hamiltonian form. A brief review of these and related concepts of classical mechanics is given in Appendix \ref{apdx:mechanics}.

VIs approximate the trajectory of a continuous-time dynamical system by discretizing its action integral
\[
L^d(\v{q}_t, \v{q}_{t+1}, h) \approx \int_t^{t+h} L(\v{q}(\tau), \dot{\v{q}}(\tau)) \d \tau
.
\]
This is a discrete-time quadrature-based approximation, denoted by $L^d$, defined by $\v{q}_t = \v{q}(t)$ and $\v{q}_{t+1} = \v{q}(t + h)$ with step size $h$.
From a Lagrangian perspective, we arrive at the discrete equations of motion
\[ \label{eqn:DEOM_L}
\pd{L^d(\v{q}_{t-1}, \v{q}_{t}, h)}{\v{q}_t} + \pd{L^d(\v{q}_t, \v{q}_{t+1}, h)}{\v{q}_t} = \v{0}
,
\]
by using a discrete analog of Hamilton's principle \cite{Marsden2001}.
Following \textcite{West2004}, \eqref{eqn:DEOM_L} can be written in position-momentum form as
\[ \label{eqn:DEOM_H}
\v{p}_t\!=\!-\pd{L^d(\v{q}_{t}, \v{q}_{t+1}, h)}{\v{q}_{t}},\!\!\!
\quad 
\v{p}_{t+1}\!=\!\pd{L^d(\v{q}_{t}, \v{q}_{t+1}, h)}{\v{q}_{t+1}},
\]
where $\v{p}_t = \partial L / \partial \dot{\v{q}}_t$ are generalized momenta.

VIs are \emph{symplectic} as they conserve phase-space volume exactly.
Symplectic integrators also approximately conserve energy, often only introducing third-order (and above) discretization error with respect to the energy.
Such integrators yield discrete-time dynamical systems that closely resemble the continuous-time systems under study, and evolve in a way that is globally consistent with the true solution.

VIs are also \emph{momentum-preserving}. This means that for any symmetry in the discrete system, there is a quantity that is exactly conserved.
These properties help to ensure their accuracy.
In the dissipative and forced cases, VIs have been both theoretically and empirically shown to produce stable long-term predictions and to capture statistically important quantities, even in chaotic regimes \cite{Lew2004}.

\section{Variational Integrator Networks}
\label{sec:vins}

To define a variational integrator network, we begin with the viewpoint of neural ODEs \cite{Haber2017, Chen2018}.
In this setting, we specify an ODE whose right-hand-side is a single-layer neural network. We then obtain a deep residual network using an Euler discretization scheme, where the depth of the network is determined by the number of discretization steps.

We mirror this viewpoint with the goal of developing network architectures that learn dynamical systems faithfully, by having their learned embeddings be dynamical systems in their own right.
Compared to neural ODEs, we introduce two key differences.
\1 Rather than constructing a free-form system of ODEs, we construct a system of ODEs arising from the Euler-Lagrange equations governing a free-form dynamical system.
\2 Instead of an Euler discretization, we use a structure-preserving discretization given by a VI.
\0

We focus on VIs with explicit discrete update equations arising from the discrete equations of motion \eqref{eqn:DEOM_L} and \eqref{eqn:DEOM_H}. This results in network architectures that do not require fixed-point algorithms to evolve the dynamics.

We begin by considering separable \emph{Newtonian networks}, i.e. networks that follow Newton's laws of physics.
These are constructed by considering a parameterized Lagrangian of the form
\[
L_{\theta}(\v{q},\dot{\v{q}}) =T_{\theta}(\dot{\v{q}}) - U_{\theta}(\v{q}) = \frac{1}{2}\dot{\v{q}}^T\m{M}_{\theta} \dot{\v{q}}  - U_{\theta}(\v{q}),
\]
where $T_{\theta}$ and $U_{\theta}$ are the kinetic and potential energy of the system, and $\m{M}_{\theta}$ is a symmetric, positive definite inertia matrix. We omit time dependence for ease of notation.
From a Lagrangian perspective, approximating the action by the quadrature rule
\[\label{eqn:sv_quadrature}
L^d(\v{q}_t, \v{q}_{t+1}, h) = hL_{\theta}\Big(\v{q}_t, \frac{(\v{q}_{t+1} - \v{q}_t)}{h}\Big)
,
\]
we arrive at the St\"{o}rmer-Verlet (SV) integrator
\[ \label{eqn:vin_layer_lagrange}
\v{q}_{t+1} = 2\v{q}_t - \v{q}_{t-1} - h^2 \m{M}_{\theta}^{-1} \pd{U_{\theta}(\v{q}_t)}{\v{q}_t}
.
\]
The symmetric variant of \eqref{eqn:sv_quadrature}, given by
\<
L^d(\v{q}_t, \v{q}_{t+1}, h) &= \frac{h}{2}\Big(L_{\theta}\Big(\v{q}_t, \frac{(\v{q}_{t+1} - \v{q}_t)}{h}\Big) 
\\
&\quad+ L_{\theta}\Big(\v{q}_{t+1}, \frac{(\v{q}_{t+1} - \v{q}_t)}{h}\Big)\Big)
,
\>
yields the velocity Verlet (VV) integrator
\<\label{eqn:vin_layer_hamiltonian}
\v{q}_{t+1} &= \v{q}_t + h\m{M}_{\theta}^{-1}\dot{\v{q}}_t - \frac{h^2}{2} \m{M}_{\theta}^{-1} \pd{U_{\theta}(\v{q}_t)}{\v{q}_t}
,
\\
\v{p}_{t+1} &= \v{p}_t - \frac{h}{2}\Big(\pd{U_{\theta}(\v{q}_t)}{\v{q}_t} + \pd{U_{\theta}(\v{q}_{t+1})}{\v{q}_{t+1}}\Big)
,
\>
where $\v{p}_t = \m{M}_{\theta}^{-1}\dot{\v{q}}_t$. Compared to \eqref{eqn:vin_layer_lagrange}, the VV integrator explicitly incorporates the momentum/velocity.

Combining variational integrators with the neural ODE viewpoint, we arrive at network architectures that enjoy the following properties.
\1 Physical properties, such as conservation laws, are automatically enforced by preserving the underlying geometric structure.
\2 Flexibility to model complex phenomena is retained, as $U_{\theta}$ can be a black-box neural network. We opt for a single-layer fully connected network.
\3 Interpretability is increased, by considering that the embedding evolves in a phase-space, having notions of kinetic and potential energy.
\4 Modeling specificity is increased, since the mass term can either be modeled explicitly or taken to be the identity matrix.
\0

\begin{figure*}
\centering
\includegraphics{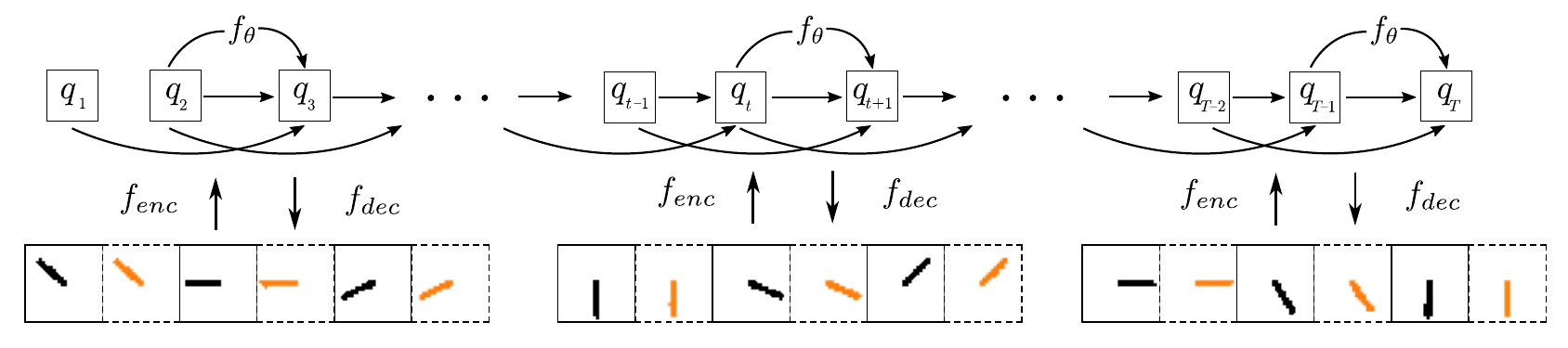}
\caption{Learning the dynamics of a pendulum from pixel observations.
Here, a variational autoencoder maps the pixels into the latent space $\v{q}$ using $f^{enc}$, and maps the latent space back into pixels using $f^{dec}$. A Lagrangian variational integrator is used, for which $\v{q}$ is the hidden state. Unlike an ordinary residual network, the skip connections used are intertwined. We display the observations in black, and predicted values given by the decoder in orange. Experimental details for this setup are given in Section \ref{sec:experiments}.} \label{fig:vae_network}
\end{figure*}

To illustrate how variational integrators enable us to build further geometric structure into the model, consider a \emph{Newtonian rotation network} in 2D.
The idea is to exploit the knowledge that a system's evolution takes place entirely on a manifold, here the space of rotations, by incorporating this structure into the network.

For this, we consider a particular class of variational integrators: \emph{Lie group variational integrators} (LGVIs).
LGVIs exploit the properties of Lie groups to construct integrators that automatically evolve on a specified Lie group.
The key idea is to approximate the change in position over integration steps using group elements \cite{Leok2007}.
Since the state space is closed under the group action (e.g. matrix multiplication when represented by matrices), the constraints are automatically enforced.
For instance, the Lie group $SO(2)$ (with matrix multiplication as the group action) is a natural way to encode the underlying manifold of 2D rotations, like the evolution of the angle of a pendulum.
A Newtonian network in a uniform gravitational potential that evolves automatically on $SO(2)$ is specified as follows.
Denoting the angle by $\vartheta$, the corresponding rotation network is given by
\<\label{eqn:vin_layer_rotation}
\sin \Delta \vartheta_t &= \sin \Delta \vartheta_{t-1} + h^2 r_\varphi(\vartheta),
\\
\vartheta_{t+1} &= \vartheta_t + \Delta \vartheta_t,
\>
where $r_{\varphi}(\vartheta_t)$ is a neural network with $\sin(\cdot)$ activations at the last layer.
Appendix \ref{apdx:so(2)} provides further details.

\subsection{Learning VINs from Noisy Observations}
Given initial conditions of a system, the state evolution is given by a solution $\v{q}(t)$ to the equations of motion. Denoting the state in phase space by $\v{x}_t=(\v{q}_{t-1}, \v{q}_t)$ from the Lagrangian perspective or $\v{x}_t=(\v{q}_t, \v{p}_{t})$ from the Hamiltonian perspective, VINs represent an approximation to the solution between the initial condition $\v{x}_1$ and terminal state $\v{x}_T$.
We represent a layer in the network by
\[\label{eqn:vin_layer_t}
\v{x}_t = f_{\theta}(\v{x}_1, h, t),
\]
as a function of the initial condition, step size and time step (layer index) $t$. 
Figure \ref{fig:vins} gives an illustration of a VIN.
Given a path of noisy observations $\v{y}_{1:T}$ of the state of a system, we specify a Gaussian likelihood
\[  \label{eqn:vin_likelihood}
p(\v{y}_{1:T}\given\v{x}_{1:T}, \sigma^2) = \prod_{t=1}^T \c{N}(\v{y}_t\given \v{x}_t, \sigma^2\m{I}).
\]
Define $\Theta = (\theta, \v{x}_1, \sigma^2)$ where $\theta$ are the parameters of the VIN, $\v{x}_1$ is the initial condition , and $\sigma^2$ is the error variance. We train the model by maximizing the log of the likelihood \eqref{eqn:vin_likelihood} with respect to $\Theta$ using stochastic optimization.

\subsection{VINs for High Dimensional Observations}
\label{sec:vin_vae}

It is possible that the dynamical system of interest is not observed directly, but indirectly through a set of intermediate data not of primary interest.
For example, we can observe a swinging pendulum by seeing images of its location at a given set of time instances.
We propose to address this problem using variational autoencoders (VAEs) \cite{Kingma2014, Rezende2014}.
VAEs enable approximate inference in latent variable models that model high-dimensional observations as being generated by some lower dimensional latent space. 
We aim to combine this setup with VINs to learn physical systems that evolve in a latent phase-space.

We start by placing a standard Gaussian $p_{\theta}(\v{x}_1) = \c{N}(\v{x}_1 \given \v{0},\m{I})$ over the initial condition.
The joint distribution over a path is
\[ \label{eqn:joint_px}
p_{\theta}(\v{x}_{1:T}) = p_{\theta}(\v{x}_1)p_{\theta}(\v{x}_{2:T}\given\v{x}_1),
\]
which we can sample from by sampling $\v{x}^s_i \dist p(\v{x}_1)$, and propagating the samples through the network $\v{x}^s_t = x_{\theta}(\v{x}^s_1, h, t)$. Assuming noise-free dynamics the uncertainty over the dynamics is fully induced by the distribution of the initial condition. 

We specify the joint distribution over observations and paths in latent space as
\[ \label{eqn:joint_vae}
p_{\theta}(\v{y}_{1:T}, \v{x}_{1:T}) = p_{\theta}(\v{x}_{1:T})\prod_{t=1}^T p_{\theta}(\v{y}_t\given\v{x}_t).
\]
The likelihood $p_{\theta}(\v{y}_t\given\v{q}_t)$ is parameterized by a decoder neural network $f^{dec}_{\theta}(\v{q}_t)$, which depends only on the position component $\v{q}_t$ of $\v{x}_t$.

We aim to approximate the posterior distribution $p_{\theta}(\v{x}_{1:T}\given\v{y}_{1:T})$, which is intractable due to the nonlinear relationships introduced by the decoder $f^{dec}_{\theta}$ and the dynamics $x_{\theta}$ in \eqref{eqn:vin_layer_t}.
In the VAE setup, we specify an approximation $q_{\phi}(\v{x}_{1:T}\given\v{y}_{1:T})$ to the posterior, parameterized by an encoder network $f^{enc}_{\phi}(\v{y}_{1:T})$, where $\phi$ are called the variational parameters. Figure \ref{fig:vae_network} illustrates the VIN-VAE setup.

We learn the parameters by variational inference.
We choose the variational family
\< \label{eqn:joint_qx}
q_{\phi}(\v{x}_{1:T}\given\v{y}_{1:T}) &= q_{\phi}(\v{x}_1)p_{\theta}(\v{x}_{2:T}\given\v{x}_1),
\\
q_{\phi}(\v{x}_1) &= \c{N}(\v{x}_1 \given \v{m}_1, \v{s}^2_1).
\>
Note that the conditional $p_{\theta}(\v{x}_{2:T}\given\v{x}_1)$ is the same in the variational family as in the model.
The mean and variance of the initial condition are in general estimated from the full trajectory $\v{y}_{1:T}$ by the encoder $f^{enc}_{\phi}$.
To train the model, we minimize Kullback-Leibler divergence with respect to the model parameters $\theta$ and the variational parameters $\phi$, which is equivalent to maximizing the evidence lower bound
\[
\sum_{t=1}^T \mathbb{E}_{\v{x}_t^s \sim q_{\phi}(\v{x}_t\given\cdot)} \log p(\v{y}_t\given\v{x}_t) \! -\! \mathbb{KL}\big[q_{\phi}(\v{x}_1) \from p(\v{x}_1)\big],
\nonumber
\]
where $\v{x}_t^s \sim q_{\phi}(\v{x}_t(\cdot))$ denotes a sample from $q_{\phi}(\v{x}_1)$ propagated through the network $x_{\theta}$, and $\mathbb{KL}\big[q \from p\big]$ denotes the Kullback-Leibler divergence between $q$ and $p$.
This objective is maximized with respect to $\theta$ and $\phi$ jointly using stochastic optimization.

\section{Experiments} \label{sec:experiments}

To study the performance of VINs, we implemented them for two reference systems: (a) an ideal pendulum, (b) an ideal mass-spring system.
We study the ability of VINs to infer a useful representation of the system when given a small quantity of data, in cases where the dynamical system is observed both directly and indirectly. Full details for network architectures and hyperparameters are given in Appendix~\ref{apdx:hyperparam}.

\subsection{Learning from Noisy Observations}

We consider VINs in a noisy setting.
Specifically, the model is given noisy position and velocity measurements from which it needs to learn the dynamics. We compare our proposed VINs with Hamiltonian neural networks (HNNs) \cite{Greydanus2019} and standard feed-forward neural networks (NNs) without additional structure that would explicitly incorporate physical or mechanical constraints.
We use the VIN given by \eqref{eqn:vin_layer_lagrange}.
HNNs are trained on observations of the form $(\v{q}_t, \v{p}_t, \dot{\v{q}}_t, \dot{\v{p}}_t)$.
We replicate the setup from \textcite{Greydanus2019} with one key difference: we introduce noise in all observations, rather than only introducing it in $(\v{q}_t, \v{p}_t)$ and observing $(\dot{\v{q}}_t, \dot{\v{p}}_t)$ noise free.
This makes the setting more realistic, but system identification harder.
To account for the noise, we add a noise variable to all models and maximize the log-likelihood, rather than only mean-squared error.

\begin{figure*}
\centering
\includegraphics[width=0.95\hsize]{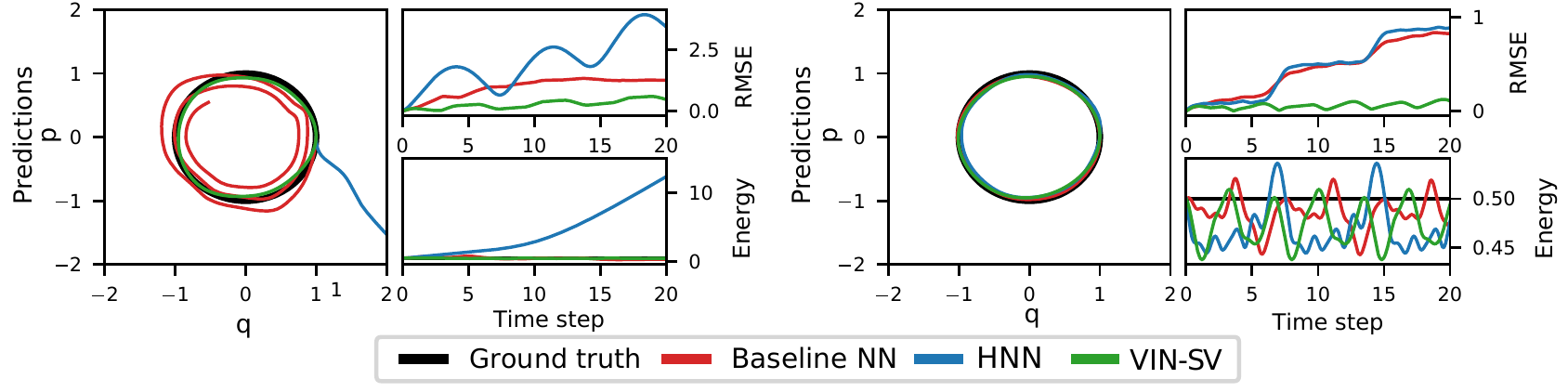}
\caption{Learning physics from noisy observations for the ideal mass-spring. Given a set of initial conditions, we forecast a path in configuration space and compare against the ground truth. We show model predictions, total root-mean-squared error between coordinates and the total energy of the dynamical system in the embedding.}\label{fig:ideal_mass_spring}
\end{figure*}
\begin{figure*}
\centering
\includegraphics[width=0.95\hsize]{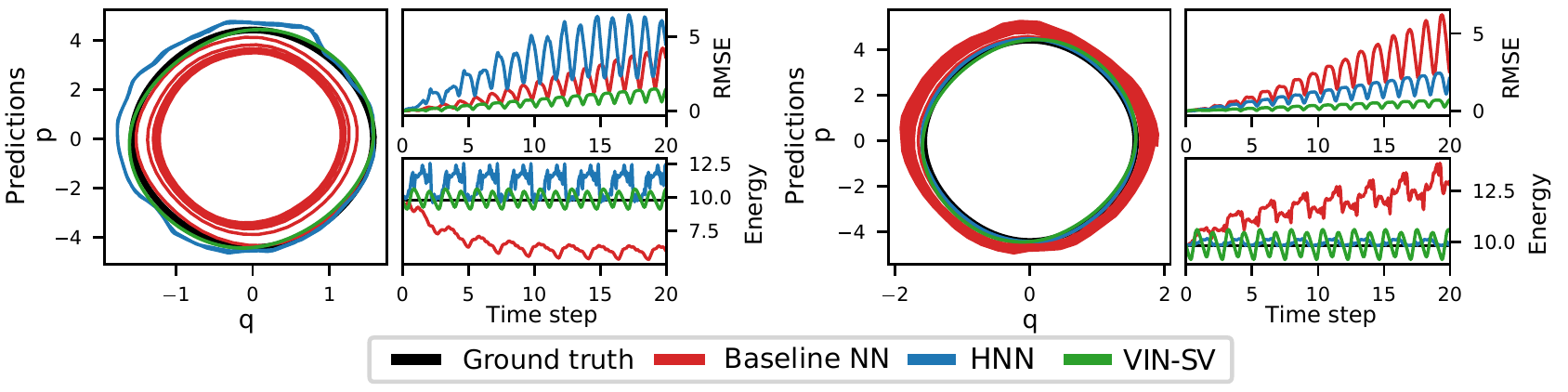}
\caption{Learning physics from noisy observations for the ideal pendulum. Given a set of initial conditions, we forecast a path in configuration space and compare against the ground truth. We show model predictions, total root-mean-squared error between coordinates and the total energy of the dynamical system in the embedding.}\label{fig:ideal_pendulum}
\end{figure*}

We examine two scenarios: (a) a moderate-data regime, where models are trained using 25 training trajectories with a total of 750 data points, (b) a low-data regime using 5 training trajectories with a total of 150 data points.
Figures \ref{fig:ideal_mass_spring} and \ref{fig:ideal_pendulum} show that prediction performance differs between the models.
In the low-data regime, despite learning to approximately conserve the system's energy, the HNN does not capture the correct dynamics, and performs poorly on prediction in terms of RMSE on both systems.
On the mass-spring system (Figure \ref{fig:ideal_mass_spring}), with sufficient data, the HNN prediction error is low over a small horizon, but exhibits two large jumps as the trajectory evolves.
We suggest that in both cases the HNN fits the noise in the training data (overfits) and fails to identify the underlying system.
The NN baseline performs better than the HNN in the low-data regime, whereas the HNN demonstrates better predictive performance in the moderate-data regime on the pendulum system (Figure \ref{fig:ideal_pendulum}).
The VIN exhibits good predictive performance, outperforming the baselines on both systems, in both the low-data and moderate-data regimes.

Figures \ref{fig:ideal_mass_spring} and \ref{fig:ideal_pendulum} show that the energy behaviors of HNNs, VINs, and NNs differ.
Given sufficient data, both the HNN and VIN learn a model that conserves a quantity that approximates the energy of the system. However, the HNN overfits in the low-data regime on both systems.
The NN baseline incorrectly dissipates/adds energy in both scenarios for the pendulum system, particularly as time passes, but learns to approximately conserve energy for the mass-spring system given 25 training trajectories.
This contributes to the worse predictive performance of the NN baseline compared to the HNN and VIN.

Overall, VINs can effectively identify the system from noisy observations, even in small-data scenarios, where HNNs and NNs can overfit.
We attribute this to their architecture: their embedded space is a dynamical system in its own right, which enforces physical constraints automatically when forecasting so that their long-term predictions better match the true system.
In contrast, the HNN relies on generalization to conserve energy, as demonstrated by the difference in performance in the low-data and moderate-data regimes.

\subsection{Learning from Pixel Observations}

\begin{figure}
\centering
\includegraphics{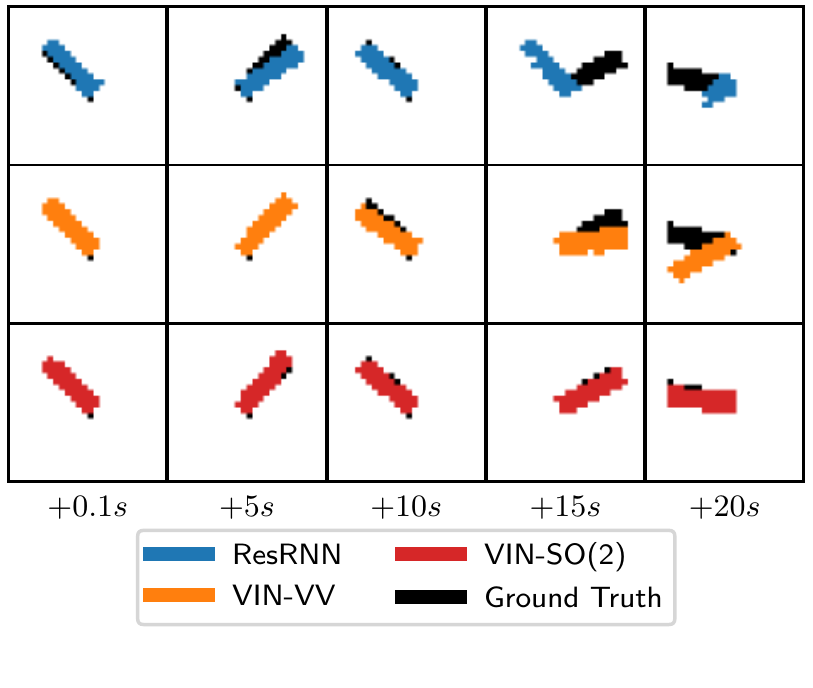}
\caption{Reconstructions of the pendulum system from forecasts in latent space, using a step-size of $0.1s$, up to $20$ seconds. The ground truth (black) is occluded by the model predictions, shown in color.}
\label{fig:pendulum_recon_forecast}
\end{figure}

\begin{figure}
\centering
\includegraphics{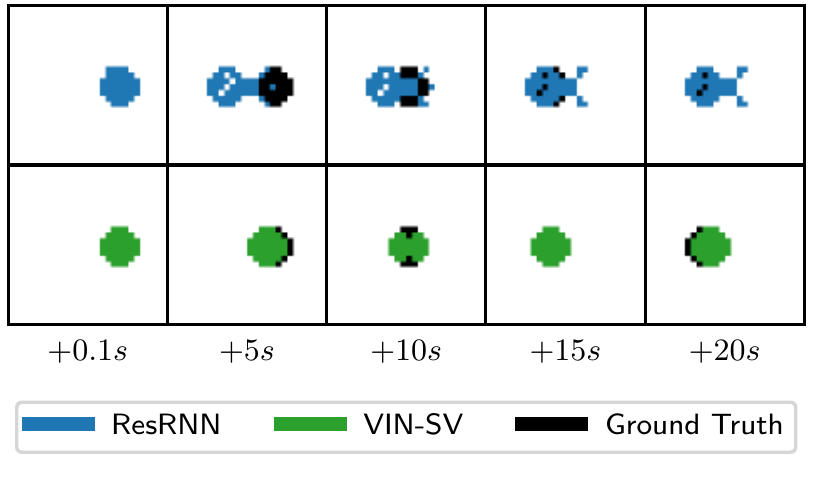}
\caption{Reconstructions of the mass-spring system from forecasts in latent space, using a step-size of $0.1s$, up to $20$ seconds. The ground truth (black) is occluded by the model predictions, shown in color. The mass oscillates left-and-right based on the initial tension in the spring (not rendered in images).}
\label{fig:springmass_recon_forecast}
\end{figure}

We study VINs in a variational auto-encoder (VAE) setting, which adds an auxiliary image processing task to prediction.
Here, we observe $28\cross 28$ pixel images depicting the mass-spring and pendulum systems; see, e.g. Figures \ref{fig:pendulum_recon_forecast}--\ref{fig:springmass_recon_forecast}.
For the mass-spring, we use \eqref{eqn:vin_layer_lagrange} for the dynamics (VIN-SV).
For the pendulum, we run experiments using both \eqref{eqn:vin_layer_hamiltonian} (VIN-VV) as well as the dynamics imposing $SO(2)$ manifold constraint in \eqref{eqn:vin_layer_rotation} (VIN-$SO(2)$).
As a baseline, we use a parameter-tied deep recurrent residual network (ResRNN) having the same number of layers as the VINs, with each layer sharing the same single-layer neural network. 
This mirrors the structure that arises from the time independence of the Lagrangian in VINs.
Each model is trained within a VAE framework as described in Section \ref{sec:vin_vae}.

\begin{figure*}
\begin{center}
\subfloat[VAE]{\includegraphics{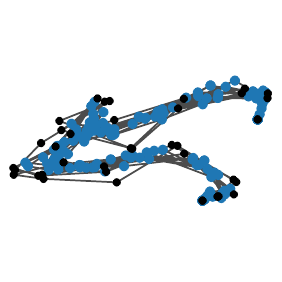}}
\hfill
\subfloat[DVAE]{\includegraphics{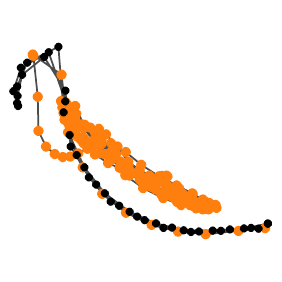}}
\hfill
\subfloat[LG-VAE]{\includegraphics{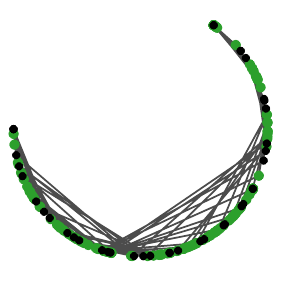}}
\hfill
\subfloat[VIN-$SO(2)$]{\includegraphics{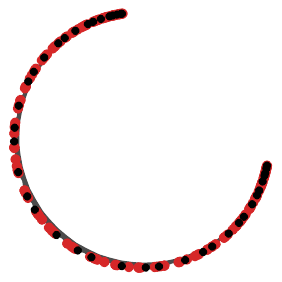}}
\hfill
\subfloat[\!VIN-$SO(2)$ \footnotesize(fixed)\!\!]{\includegraphics{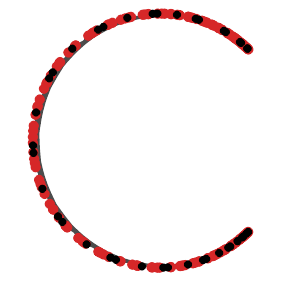}}
\hfill
\subfloat[Ground Truth]{\includegraphics{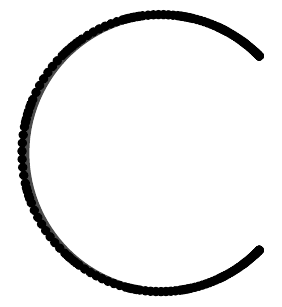}}
\end{center}
\caption{Example embedded representations of an ideal pendulum system. Black/colored dots represent embedded train/test images, gray lines connect points sequentially in time. The embeddings learned by the baseline models fail to capture the global structure (a)--(b) and/or are discontinuous with respect to the time dimension (c). The VIN-$SO(2)$ (d)--(e) learns an embedding that is consistent with the ground truth (f). In (e), we fix the mass matrix, which is not identifiable from pixel data, to the true value. Here, the VIN-$SO(2)$ faithfully reconstructs the ground truth.}
\label{fig:free_pendulum_representations}
\end{figure*}

We evaluate the structure of the latent space learned by VIN-$SO(2)$ and compare it with representations learned by a standard VAE \cite{Kingma2014, Rezende2014}, a VAE with free-form dynamics governed by a feed-forward network (DVAE), and a Lie group VAE (LG-VAE) \cite{Falorsi2018} with no dynamic structure.
Figure \ref{fig:free_pendulum_representations} visualizes the latent spaces after training on $4s$ (40 observations) and mapping an additional $80$ test images into latent space using the encoder $f^{enc}_{\phi}$, including the dynamics in the case of the VIN-$SO(2)$.
The VAE captures local structure: observations close together in image space are mapped to points close together in latent space.
However, it fails to capture the global structure of the state space and has discontinuities with respect to the sequential nature of the dataset.
Figure \ref{fig:free_pendulum_representations}(b) shows that adding an unrestricted neural network to capture the dynamics does not solve the problem.
The LG-VAE captures the correct global structure by restricting the manifold, but still exhibits discontinuities with respect to the time dimension, since it does not model the dynamics.
The embedding for VIN-$SO(2)$ does not have such discontinuities: it learns both the global structure and respects the sequential nature of the data due to the structure encoded by the VIN.

\begin{figure}
\includegraphics[width=\hsize]{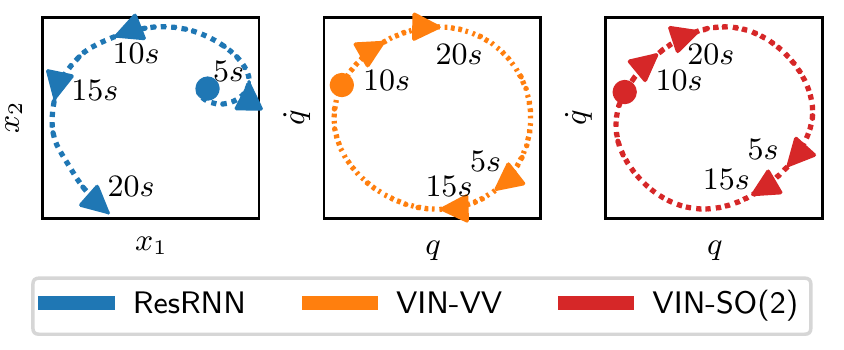}
\caption{Latent embeddings learned from pixel observations of an ideal pendulum. Circles denote the inferred initial condition, dots denote predictions forward in time. Triangles mark $5$-second intervals in the forecasts. The ResRNN fails to capture the underlying geometric structure and spirals far beyond the initial condition.  VINs preserve this structure automatically.}
\label{fig:latent_pendulum_forecast}
\end{figure}

For both systems, we generate $6$ seconds of training observations, sampled at a frequency of $10$Hz (60 observations). Training data is split into overlapping $1s$ image trajectories ($10$ observations), matching the depth of the networks, which was $10$ in all experiments.

We assess the models qualitatively by looking at the properties of their latent spaces. In particular, we infer a distribution over the initial condition using the learned encoder $f^{enc}_{\phi}$ given $10$ initial observations from the pendulum system. We then evolve the learned system for $20$ seconds using the mean of the variational posterior.

Figure \ref{fig:latent_pendulum_forecast} shows how the ResRNN does not learn dynamics that match the geometric properties of the true system (i.e. symplectic) but instead spirals away from the initial condition (denoted by the large circle). This is because the Euler discretization scheme used by residual networks ignores the underlying geometry.  On the other hand, both the VIN-VV and the VIN-$SO(2)$ models automatically preserve symplectic structure and evolve strictly on a sub-manifold in their respective latent phase-spaces. Importantly, while the flexibility afforded by the decoder allows the ResRNN setup to generate plausible observations up to some fixed horizon, the unbounded behavior of the evolution eventually causes significant failures.

Figure \ref{fig:pendulum_recon_forecast} shows the reconstructions obtained by mapping the latent paths from Figure \ref{fig:latent_pendulum_forecast} through the decoder $f^{dec}_{\theta}$. Between $10s$--$15s$ of forecasting, the ResRNN predictions are  unreliable: going through discontinuous jumps in pixel space, suddenly reversing the dynamics and generating half-formed pendula (see, e.g. the final step in Figure \ref{fig:latent_pendulum_forecast}).

Conversely, the VINs do not exhibit such non-physical behavior, since the latent path remains bounded on the data manifold despite forecasting for effectively arbitrary long horizons. The VIN-VV does display signs of going out of phase with the ground truth around $15s$ in Figure \ref{fig:pendulum_recon_forecast}, becoming more pronounced around the $20s$ mark. One explanation is that we only consider the path traversed by the mean of the variational posterior, and ignore the build-up in uncertainty as the prediction horizon increases. However, looking at the same reconstructions from the VIN-$SO(2)$ model, we see that it does not suffer from this problem within the $20s$ prediction horizon. Therefore, we assume that the error from assuming an Euclidean manifold contributes to the mismatch as well.

We perform the same qualitative analysis on reconstructions of the mass-spring system, shown in Figure \ref{fig:springmass_recon_forecast}. Although the underlying system is simpler in this instance, the performance of the ResRNN deteriorates even quicker with increasing prediction horizon. The VIN-SV also exhibits small errors in the reconstruction at the $10s$ mark, but captures the underlying dynamics well, as can be seen by its long-term predictions.

\begin{table}
\begin{center}
\scalebox{0.9}{
\begin{tabular}{l l c c}
\textbf{System} & \textbf{Model}  & RMSE & $\!\log p(\v{y}\given\v{x}) \times 10^2\!$ \\
\hline 
& ResRNN & $6.1 \pm 0.2$ & $-246.7 \pm 79.2$ \\
Pendulum\!\! & VIN-VV & $4.3 \pm 0.6$ & $-13.4 \pm 5.8$ \\
& VIN-$SO(2)$\!\!\! & $\mathbf{3.4 \pm 0.6}$ & $\mathbf{-3.2 \pm 1.9}$ \\
\hline 
Mass- & ResRNN & $6.1 \pm 0.1$ & $-4.7 \pm 2.4$ \\
spring & VIN-SV & $\mathbf{3.2 \pm 0.2}$  & $\mathbf{-0.2 \pm 0.0}$
\end{tabular}
}
\end{center}
\caption{RMSE and log-likelihood (with standard errors) for the pendulum and mass-spring systems over $5s$ forecasts on pixel observations.} \label{tbl:rmse}
\end{table}

We perform a quantitative analysis with a similar setup on both systems. Specifically, we run $10$ randomized trials, where we generated $6$ seconds of observations to train on and use the same architectures as before. In each trial, we then infer a distribution for the initial condition on the same trajectory and evaluated the RMSE and log-likelihood for $5s$ forecasts. We evaluate on the training trajectory to isolate properties of the dynamics, which is only trained on $1s$ forecasts (i.e. having 10 layers). Table \ref{tbl:rmse} shows the results with standard errors. Both VINs perform significantly better than the ResRNN in terms of both RMSE and log-likelihood. The VIN-$SO(2)$ shows a meaningful improvement in terms of log-likelihood when compared to the VIN-VV, whereas the RMSE is inconclusive.

\section{Discussion}

VINs can be used to create embeddings that faithfully represent dynamical systems.
This enables them to learn with less data and provides greater interpretability compared to other network architectures, while facilitating accurate long-term predictions.
Provided their state space is chosen appropriately, VINs preserve the topological and geometric structure of the dynamical systems they encode.
This assists with performance, mirrors recent developments in VAEs designed to accurately encode physical systems \cite{Gong2018, Haber2017, Lutter2019, Caterini2018}, and is well-motivated by recent theoretical observations made in the context of neural ODEs \cite{Dupont2019}.

The imposition of additional geometric structure does not cause VINs to lose their capacity to model flexible classes of phenomena.
In particular, they are still parameterized by an underlying neural network.
This mirrors the design of residual networks and other architectures related to differential equations \cite{Haber2017, Chen2018}.
Thus, VINs are more interpretable than purely black-box approaches to network design, while still being highly expressive.
VINs can be trained directly on noisy observations.
They may also be used as part of larger and more complex learning pipelines, e.g. by incorporating them into an auto-encoding framework.
Performance in both settings is discussed in Section \ref{sec:experiments}.

A number of directions could be pursued to improve these ideas.
In particular, one could study these ideas with time-varying Lagrangians, improving expressivity by greatly expanding the class of dynamical systems faithfully representable by the embedding.
This would bring VINs closer in line with residual networks and general neural ODEs \cite{Chen2018}.
While we focused on data efficiency and representation learning in settings where the underlying dynamics are fairly simple, it would be interesting to study such networks on more complex tasks.
This could pave the way toward better performance on currently difficult problems in areas where the phenomena under study are dynamical systems, such as robotics and reinforcement learning.

\section{Conclusion}

In this work, we introduced \emph{variational integrator networks}, a class of deep network architectures for creating neural embeddings, which encode and represent dynamical systems.
VINs ensure faithful representation of dynamical systems by using an embedding that forms a dynamical system in its own right.
This facilitates data-efficient learning, enhances interpretability, and allows for accurate long-term predictions when compared to other classes of networks.

Recent trends in deep learning have sought to improve the performance of deep networks on physical systems by designing networks whose behavior is more understandable and better matched to the underlying physics.
Variational integrator networks take a step toward progressing this line of work.

\section*{Acknowledgments}

This work was supported by Microsoft Research through its PhD scholarship program.

\printbibliography

\newpage
\onecolumn
\appendix

\section{Appendix: Short Review of Lagrangian and Hamiltonian Mechanics} \label{apdx:mechanics}

Hamiltonian and Lagrangian mechanics are two intricately related formulations of classical mechanics.
In classical mechanics, we assume that we are given a continuous-time dynamical system defined on a space $\c{Q} \subseteq \R^d$, which we call the \emph{configuration space}.
A state of the system is taken to be a set of parameters $\v{q}\in\c{Q}$ that uniquely identify the configuration of the system.
Continuous-time evolution of the dynamics in $\c{Q}$ yields a path in configuration space.
Lagrangian and Hamiltonian mechanics formulate the laws of physics in terms of properties of these paths.

Specifically, \emph{Hamilton's principle}, also called the \emph{Principle of Least Action}, states that there exists a real-valued function $L$ such that all paths in configuration space which occur in nature minimize the path integral
\[
S(\v{q}) = \int_0^T L(\v{q}(t), \dot{\v{q}}(t)) \d t
\]
where $\dot{\v{q}}$ is the velocity, which is the time-derivative of position.
For a given $L$, it can be shown using the calculus of variations that minimization of $A$ is equivalent to solving a system of partial differential equations
\[
\od{}{t}\Big(\pd{L}{\dot{\v{q}^q}}\Big) - \pd{L}{{\v{q}^q}} = \v{0},
\]
called the \emph{Euler-Lagrange Equations}, or the \emph{equations of motion}.
Given a set of initial conditions $(\v{q}(0), \dot{\v{q}}(0))$, the solutions to the equations of motion describe the trajectory of the system.

This gives the starting point of Lagrangian mechanics -- physical phenomena that satisfy it are called \emph{classical}, and span virtually all areas of physics.
The behavior of particular phenomena varies according to choice of the Lagrangian $L$, which fully characterizes how the system evolves over time.

For example, for $\v{q}\in\R^d$, take $L(\v{q}, \dot{\v{q}}) = T(\v{q}, \dot{\v{q}}) - U(\v{q})$ where $T$ is the kinetic energy, and $U$ is the potential energy of the system.
This describes a conservative Newtonian system.

\section{Appendix: Lie Group Variational Integrator for $SO(2)$} \label{apdx:so(2)}

We start by formulating a Lagrangian with the Lie group $SO(2)$ using matrix representations. First, define the map from scalars $\omega \in \mathbb{R}$ to $2\times 2$ skew-symmetric matrices
\[
\m{S}(\omega) =
\begin{bmatrix}
0 & -\omega \\
\omega & 0
\end{bmatrix}.
\]
The set of $2\times 2$ skew-symmetric matrices forms the Lie algebra $\fr{so}(2)$. The matrix exponential map, takes elements of the Lie algebra to elements of the group $SO(2)$
\[
\m{R}(\omega) = \exp \m{S}(\omega) =
\begin{bmatrix}
    \cos \omega & -\sin \omega \\
    \sin \omega & \cos \omega
\end{bmatrix}.
\]
Kinematics for group elements $R \in SO(2)$ can be written in terms of Lie algebra elements as
\[
\dot{\m{R}} = \m{R}\m{S}(\omega),
\]
where $\omega$ is analogous to angular velocity. A conservative Newtonian Lagrangian in a uniform gravitational potential can be written in terms of the Lie group $SO(2)$ as
\[
L(\m{R}, \m{S}(\omega)) = \frac{1}{2}ml^2\omega^2+ mgl\v{e}_2^T\m{R}\v{e}_1
\]
where $\m{R} = \m{R}(\theta)$ is a rotation matrix parameterized by $\theta$, $g$ is the gravitational acceleration and $\v{e}_1, \v{e}_2$ are orthogonal unit vectors in the inertial frame of reference, $\v{e}_1 = [1, 0], \v{e}_2 = [0, 1]$.

To develop a Lie group variational integrator, define $\m{F}_t \in SO(2)$ such that
\[ \label{eqn:lgvi_update}
\m{R}_{t+1} = \m{R}_t\m{F}_t.
\]
Since $\m{F}_t \in SO(2)$, the update enforces $\m{R}_{t+1} \in SO(2)$ since Lie groups are closed under the group action.
Here, group action is given by matrix multiplication.
Then define the discretization of the action integral as
\[
L^d(\m{R}_k, \m{F}_k) = \frac{1}{2h}ml^2 \langle \m{F}_k - \m{I}, \m{F}_k - \m{I} \rangle + \frac{hmgl}{2}\big(\v{e}_2^T\m{R}_t\v{e}_1 + \v{e}_2^T\m{R}_{t+1}\v{e}_1\big),
\]
which approximates the angular velocity as
\[
\m{S}(\dot{\theta}) = \frac{\m{F}_k - \m{I}}{h}.
\]
Using the discrete form of Hamilton's principle, one obtains \cite{Meyers2009} the equation
\[ \label{eqn:lgvi}
(\m{F}_t - \m{F}_t^T) - (\m{F}_{t+1} - \m{F}_{t+1}^T) - \frac{2h^2g}{l}\m{S}(\v{e}_2^T\m{R}_{t+1}\v{e}_1) = \v{0},
\]
which, when taken with \eqref{eqn:lgvi_update}, defines the Lie group variational integrator.
One arrives at \eqref{eqn:vin_layer_rotation}, written in terms of the elements of the matrices, by subsuming the force terms into the neural network.

\section{Appendix: Hyperparameters for Experiments} \label{apdx:hyperparam}

\subsection{Noisy System Observations}

The setup resembles the one of \textcite{Greydanus2019} closely.
The neural network architecture for the baseline NN, the network that parameterizes the Hamiltonian in HNNs and the one that parameterizes the VIN was the same throughout.
This was a single hidden layer feed-forward network with $200$ hidden units and $\tanh(\cdot)$ activations on the hidden layer.
The noise added to the observations was sampled from a standard Gaussian with standard deviation $\sigma=0.1$.
For the mass-spring system, we set the spring constant and mass to $k = m = 1$, as was done by \textcite{Greydanus2019}.
For the pendulum, unlike the original work, we use $m=l=1$, and $g=9.81$.
Training trajectories were sampled uniformly from energies ranging from $[0.2, 1]$ for the mass-spring system and $[1.3, 2.3]$ for the pendulum. We trained the models using ADAM with a learning rate of $10^{-3}$.
We did a hyperparameter search over $[2000, 5000, 10000]$ training steps and chose the best performing models for comparison.

For predictions with the baseline NN and HNN, we use the procedure of \textcite{Greydanus2019}, which uses fourth order Runga-Kutta with an error tolerance of $10^{-9}$, implemented in \textsc{scipy.integrate.solve\_ivp}.
For the VIN we simply predict forwards in time using the trained network.

\subsection{Pixel Observations}
In all VAE experiments we used the same encoder and decoder structure. Both the encoder and decoder consisted of two fully connected hidden layers with a $1000$ hidden units and ReLU activation functions.
\1* Encoder: two fully-connected hidden layers with $1000$ units and ReLU activation functions, followed by an LSTM with a $50$ dimensional hidden state that processed the embedded sequence in reverse to give the variational parameters for the initial condition.
\2* Decoder: two fully-connected hidden layers with $1000$ units and ReLU activation functions.
\0*

The dynamics networks (i.e. ResRNN, VIN-VV, VIN-$SO(2)$, VIN-SV) all had a depth of $10$ and used $10$ observations as input to the encoder. The step size for the networks was chosen to be $1.0$ in latent space. The underlying fully connected network had $1000$ hidden units and $\tanh$ activation functions.

We train using ADAM with a learning rate of $3.0\times 10^{-4}$ until the ELBO converges on the training set, up to a maximum of $6000$ epochs through the datasets and use the parameters with the highest ELBO for evaluation.

\end{document}